\documentclass[11pt,a4paper]{article}
\usepackage[utf8]{inputenc}
\usepackage{cmap}          
\usepackage[T1]{fontenc}
\usepackage[margin=2.5cm]{geometry}
\usepackage{amsmath}
\usepackage{amssymb}
\usepackage{graphicx}
\usepackage{booktabs}
\usepackage{array}
\usepackage{enumitem}
\usepackage{hyperref}
\usepackage{titlesec}
\usepackage{parskip}
\usepackage[expansion=false]{microtype}
\DisableLigatures{encoding = *, family = *}

\hypersetup{
    colorlinks=true,
    linkcolor=black,
    urlcolor=blue,
    citecolor=black,
}

\titleformat{\section}{\large\bfseries}{\thesection}{1em}{}
\titleformat{\subsection}{\normalsize\bfseries}{\thesubsection}{1em}{}

\title{\textbf{Inter-Rater Reliability of LLM and Rule-Based Annotation} \\
\textbf{for Inferential Narrative Features:} \\
\large Three Studies on a Turkish Corpus \\
\normalsize Version 1.0}

\author{Levent Bulut \\
Independent Researcher \\
ORCID: \href{https://orcid.org/0009-0007-7500-2261}{0009-0007-7500-2261} \\
\texttt{levent@leventbulut.com} \, $\vert$ \, \href{https://leventbulut.com}{leventbulut.com}}

\date{August 2026 \\[4pt]
\small Document type: Empirical reliability report (three studies) \\
Framework: Bulut Doctrine / Objective Projection}

\begin{document}
\maketitle

\begin{center}
\fbox{\parbox{0.92\textwidth}{\small
\textbf{Declaration of conflicts of interest.}
Three conflicts affect this report and the reader should weigh them before the results.
\emph{First}, the author designed the six annotation rules that are evaluated here, and served as the sole human rater in Study 1. A scheme's designer is not a neutral judge of whether the scheme is learnable.
\emph{Second}, one of the evaluated systems \textemdash{} Claude (Anthropic) \textemdash{} also assisted in preparing the analysis scripts and this manuscript. A model family that is being scored cannot be treated as a disinterested analyst of its own score, and the reader should assume the framing of Section 6 is affected. The scoring itself is deterministic and reproducible from published files, so this conflict touches interpretation, not arithmetic.
\emph{Third}, the corpus under evaluation is the author's own published dataset, and a finding of low reliability reduces the evidential standing of that dataset's shipped annotations. It is reported anyway.
}}
\end{center}

\section*{Abstract}

Datasets that ship automatically generated feature annotations invite a question that is rarely asked of them: would a human agree with those labels? This report answers that question for the Objective Projection corpus, a Turkish narrative dataset whose scenes carry a per-scene \texttt{applied\_rules} field produced by a rule-based detector over six craft features \textemdash{} two prohibitions (explicit emotion labelling, simile) and four positive techniques (materialized metaphor, micro-focus, temporal anchor, atmosphere contradiction).

Three studies are reported. Study 1 ($n = 120$) scores the detector against blind labels produced by the scheme's own author. Study 2 ($n = 100$, a disjoint scene set) scores the detector plus Gemini 2.5 Flash and Grok against an independent non-expert human rater whose labels were locked before any machine ran. Study 2b re-runs the identical protocol with Claude Fable 5 (High) and ChatGPT 5.5.

The central result concerns one rule. On materialized metaphor \textemdash{} the feature closest to the methodology's theoretical core \textemdash{} the five machine labellers returned positive rates of $0$, $1$, $40$, $72$ and $78$ out of $100$ scenes, against a human count of $9$. Cohen's $\kappa$ was at or indistinguishable from chance for five of the six machine labellers, across both human references, on both scene sets: $0.004$, $0.015$, $0.000$, $0.019$, $0.027$. Raw agreement, by contrast, ranged from $74.7\%$ to $84.5\%$, an artefact of class imbalance rather than a sign of competence.

We deliberately do not resolve the finding into a single story. Two readings survive the data: that the feature is genuinely inferential and beyond current automatic detection, or that the rule's definition is not yet operational enough for \emph{any} rater to apply consistently \textemdash{} including the human. Distinguishing them requires a second independent human rater, which this report does not have and therefore does not claim.

\textbf{Keywords:} inter-rater reliability, Cohen's kappa, annotation quality, class imbalance, LLM-as-annotator, computational narratology, show-don't-tell, Turkish corpus, Bulut Doctrine

\section{Introduction}

\subsection{The problem with shipped annotations}
A growing number of published corpora carry machine-generated annotation layers: sentiment scores, discourse labels, stylistic flags. These layers are convenient. They are also, in most cases, unvalidated \textemdash{} a gap that has been argued to require task-by-task validation against human labels rather than trust in general model capability \cite{pangakis2023} \textemdash{} their reliability against human judgement is neither measured nor reported, and downstream users inherit them as if they were ground truth.

The corpus examined here is one such case, and it is the author's own. The Objective Projection dataset \cite{bulut2026dataset} publishes 500 annotated Turkish\textendash{}English scene pairs; since version 7, each scene carries an \texttt{applied\_rules} field generated by \texttt{apply\_rules.py}, a bilingual rule-based heuristic that flags six craft features. Those flags have been used to describe the corpus, to select examples, and to characterise sub-patterns. Whether a human would agree with them was, until these studies, unknown. An earlier report in this line applied the same posture to a different construct, operationalizing it on real text and publishing the divergence rather than adjusting the formula \cite{bulut2026sn}; this report extends that posture from a proposed measure to the corpus's own annotation layer.

\subsection{The six features}
The scheme is drawn from the Objective Projection methodology \cite{bulut2026method}, which sits within a broader architectural framework \cite{bulut2026framework}. It distinguishes two \emph{constitutional} rules \textemdash{} prohibitions the text must not violate \textemdash{} from four \emph{working} rules, which are positive techniques the text may deploy.

\begin{description}[leftmargin=1.4cm,style=nextline]
\item[Emotion embargo] The narrator's voice must not name an emotion directly (\emph{he was sad}, \emph{a great fear}). Physical action that a reader may interpret emotionally (\emph{he wept}, \emph{she shook}) is not a violation. Dialogue is exempt: a character may say anything.
\item[Simile prohibition] No overt simile markers (Turkish \emph{gibi}, \emph{sanki}, \emph{adeta}).
\item[Materialized metaphor] An abstract inner state is rendered as a concrete, measurable physical detail instead of being named.
\item[Micro-focus] The narration concentrates on a narrow, small, concrete object or detail.
\item[Temporal anchor] A concrete time, duration or measurement is present (a clock time, \emph{three days}, \emph{14\,\textdegree C}, \emph{two metres}).
\item[Atmosphere contradiction] The scene contains an unexpected detail that runs against its prevailing situation.
\end{description}

The emotion embargo rests on a distinction the methodology draws between naming an emotional state and rendering it as physical output \cite{bulut2026biophys, bulut2026cortical}; materialized metaphor is the positive counterpart of that same distinction, and its relation to Eliot's objective correlative is treated separately \cite{bulut2026correlative, bulut2026eliot}.

The first two are surface features: a string search can approximate them. The last four require judgement, and materialized metaphor requires the most \textemdash{} it asks the rater to decide whether a physical detail is \emph{doing} representational work, which is not a property of the string.

\subsection{What this report claims and does not claim}
It reports agreement statistics. It does not claim that any of the six features is a valid construct, that the corpus is well built, or that the detector should be trusted or discarded. It measures one thing \textemdash{} whether independent raters, human and machine, assign the same labels to the same scenes \textemdash{} and reports the answer, including where the answer is uninformative.

\section{Method}

\subsection{Design common to all three studies}
Every study compares a set of binary labels (feature \textsc{present} / \textsc{absent}) from one or more machine labellers against a human reference on the same scenes. The human reference is fixed before any machine output is inspected. Scoring is deterministic: Cohen's $\kappa$ \cite{cohen1960}, raw agreement, and the confusion counts TP/FP/FN/TN with \textsc{present} as the positive class and the human as reference.

The interpretation of such coefficients in corpus annotation is itself contested, and small differences in their values do not translate straightforwardly into differences in annotation quality \cite{artstein2008}.

Where expected agreement equals 1 \textemdash{} both raters assign the same label to every scene \textemdash{} $\kappa$ is undefined and reported as \emph{n/a} rather than as $0$ or $1$.

\subsection{Study 1: the detector against the scheme's author ($n=120$)}
One hundred and twenty Turkish scenes were labelled by a single blind human rater. That rater was the author of the methodology. This is a limitation of the first order and is discussed in Section 7; it is stated here because it conditions everything that follows.

During the first pass the rater observed that his own criterion for materialized metaphor was drifting \textemdash{} earlier and later scenes were being judged against different implicit standards. The rule was therefore re-labelled in a separate corrected pass, and the corrected labels are the ones used. The drift is recorded in the published label file rather than concealed.

\subsection{Study 2: three machine labellers against an independent rater ($n=100$)}
One hundred Turkish scenes were drawn (selection seed 2026) from a 180-scene pool untouched by Study 1, then presentation-shuffled (seed 2027) and stripped of their corpus identifiers. \textbf{The Study 1 and Study 2 scene sets do not overlap at all.}

The human rater for Study 2 was a non-expert volunteer with no involvement in the methodology, working blind to all machine output. The labels were locked \textemdash{} committed to a file marked as immutable \textemdash{} before any machine was run.

The term \emph{held-out} is used in this report in a restricted sense: the scenes are held out from rater and detector development, not from the corpus. They are corpus scenes, and the mapping from masked to corpus identifiers is published.

Three machine labellers were scored: the rule-based detector, Gemini 2.5 Flash, and Grok. The two language models received the scenes in ten prompt blocks of ten scenes each, with identical instructions and definitions in every block.

\subsection{Study 2b: two further models, identical protocol}
Study 2b re-ran the Study 2 protocol without modification \textemdash{} same 100 scenes, same locked human reference, same ten prompt blocks \textemdash{} with Claude Fable 5 (High) and ChatGPT 5.5, both accessed through their public web interfaces in fresh sessions with no project context and no memory.

Two procedural rules were fixed in advance. \emph{First-run-counts}: the first response to each block is the datum; re-runs are recorded but do not replace it. \emph{Degenerate-output rejection}: an output showing a mechanical repeating pattern is discarded rather than scored. One model, Gemini 3.6, returned thirty identical label rows and was rejected under this rule; it does not appear in the results.

Block re-runs were used to measure within-model stability: ChatGPT reproduced 58 of 60 re-run labels, Claude 60 of 60.

\subsection{Reproducibility}
All human reference labels, all model label files, the detector labels, the ten prompt blocks, the identifier mapping and the scoring scripts are published in the \texttt{evaluation/} directory of the dataset \cite{bulut2026dataset}. Every figure in Sections 3\textendash{}5 was recomputed from those files in preparing this report. Two provenance caveats apply and are stated in Section 7.

\section{Study 1: the detector against the scheme's author}

Table \ref{tab:study1} gives the full result.

\begin{table}[htbp]
\centering
\small
\begin{tabular}{lrrrrrrrr}
\toprule
Rule & Human + & Det.\ + & TP & FP & FN & Prec. & Rec. & $\kappa$ \\
\midrule
Emotion label        &   2 &  12 &   2 & 10 &  0 & 0.167 & 1.000 & \phantom{-}0.265 \\
Simile               &   2 &   2 &   2 &  0 &  0 & 1.000 & 1.000 & \phantom{-}1.000 \\
Materialized metaphor&  63 &  95 &  50 & 45 & 13 & 0.526 & 0.794 & \phantom{-}0.004 \\
Micro-focus          & 118 &  96 &  94 &  2 & 24 & 0.979 & 0.797 & -0.032 \\
Temporal anchor      & 120 & 100 & 100 &  0 & 20 & 1.000 & 0.833 & \phantom{-}0.000 \\
Atmosphere contrad.  &  12 &  17 &   5 & 12 &  7 & 0.294 & 0.417 & \phantom{-}0.258 \\
\bottomrule
\end{tabular}
\caption{Study 1 ($n=120$). Detector against a single blind human rater (the methodology's author). Overall raw agreement across all 720 cells: $81.5\%$.}
\label{tab:study1}
\end{table}

Three observations.

\textbf{The surface rules behave.} Simile is detected perfectly ($\kappa = 1.00$): both rater and detector found the same two instances. This is the expected result for a feature that reduces to a short list of Turkish function words, and it serves as a positive control \textemdash{} the pipeline is capable of agreement when the feature is a string.

\textbf{The emotion rule over-triggers.} The detector flagged twelve scenes; the human found two. Recall is perfect and precision is $0.17$. The detector is not missing violations, it is inventing them \textemdash{} almost certainly by treating observed physical action (\emph{wept}, \emph{shook}) as emotion naming, which the rule explicitly exempts.

\textbf{Materialized metaphor is at chance.} $\kappa = 0.004$. Raw agreement is $51.7\%$ \textemdash{} worse than the other rules precisely because this is the only rule in Study 1 with a near-balanced human distribution (63 present, 57 absent), so raw agreement has nowhere to hide. The detector says \textsc{present} for 95 of 120 scenes; the human says 63. They coincide often enough to look like partial success and disagree in a pattern indistinguishable from independent coin flips.

Micro-focus and temporal anchor show high precision and high raw agreement with $\kappa$ at or below zero. This is not a paradox: the human marked 118 of 120 and 120 of 120 scenes positive respectively. When one rater says \textsc{present} to everything, $\kappa$ has no variance to work with and the statistic is uninformative. The correct reading is that these two rules were \emph{not tested} in Study 1, not that they failed.

\section{Study 2: three machine labellers against an independent rater}

The independent rater's distribution across the 100 scenes was: emotion label 0, simile 1, materialized metaphor 9, micro-focus 96, temporal anchor 99, atmosphere contradiction 44. Only atmosphere contradiction falls in a distribution that supports informative $\kappa$.

\begin{table}[htbp]
\centering
\small
\resizebox{\textwidth}{!}{%
\begin{tabular}{lrrrrrrrrrr}
\toprule
 & H & \multicolumn{3}{c}{Detector} & \multicolumn{3}{c}{Gemini 2.5 Flash} & \multicolumn{3}{c}{Grok} \\
\cmidrule(lr){3-5}\cmidrule(lr){6-8}\cmidrule(lr){9-11}
Rule & + & + & TP/FP/FN & $\kappa$ & + & TP/FP/FN & $\kappa$ & + & TP/FP/FN & $\kappa$ \\
\midrule
Emotion label         &  0 &   0 & 0/0/0    & n/a              &  0 & 0/0/0   & n/a              &  0 & 0/0/0   & n/a \\
Simile                &  1 &   0 & 0/0/1    & \phantom{-}0.000 &  0 & 0/0/1   & \phantom{-}0.000 &  0 & 0/0/1   & \phantom{-}0.000 \\
Materialized metaphor &  9 &  72 & 7/65/2   & \phantom{-}0.015 &  1 & 1/0/8   & \phantom{-}0.185 &  0 & 0/0/9   & \phantom{-}0.000 \\
Micro-focus           & 96 &  81 & 78/3/18  & \phantom{-}0.022 &  9 & 9/0/87  & \phantom{-}0.008 & 82 & 78/4/18 & -0.070 \\
Temporal anchor       & 99 &  82 & 81/1/18  & -0.019           & 93 & 92/1/7  & -0.018           & 95 & 94/1/5  & -0.017 \\
Atmosphere contrad.   & 44 &   0 & 0/0/44   & \phantom{-}0.000 &  2 & 2/0/42  & \phantom{-}0.051 &  6 & 3/3/41  & \phantom{-}0.020 \\
\midrule
Overall raw           &    & \multicolumn{3}{c}{74.7\%} & \multicolumn{3}{c}{75.7\%} & \multicolumn{3}{c}{86.3\%} \\
\bottomrule
\end{tabular}%
}
\caption{Study 2 ($n=100$). Three machine labellers against an independent blind human rater. ``H\,+'' is the human positive count; ``+'' is each labeller's positive count. Detector figures were recomputed from published label files. Gemini and Grok confusion counts were recovered arithmetically from the reported $\kappa$ and agreement values (Section 4.1); their per-scene labels are not available. Agreement between Gemini and Grok, ignoring the human, was $85.7\%$.}
\label{tab:study2}
\end{table}

\subsection{Recovering the Gemini and Grok confusion counts}
\label{sec:recovery}
The per-scene label files for Gemini and Grok were lost. Their per-rule $\kappa$ and raw agreement values survive in the original analysis, and these are sufficient to recover the full confusion counts without the labels.

For a binary rule over $n=100$ scenes with a known human positive count $H$ and a known raw agreement count $A = TP + TN$, the entire table is a function of a single unknown:
\[
TN = A - TP, \qquad FN = H - TP, \qquad FP = n - A - H + TP .
\]
Cohen's $\kappa$ then fixes $TP$. Enumerating $TP \in [0, H]$ and retaining only solutions whose $\kappa$ rounds to the reported value yielded a unique solution for eleven of the twelve rule-by-labeller cells. The twelfth \textemdash{} Grok on atmosphere contradiction \textemdash{} admitted two solutions, $TP=3$ and $TP=4$; the original write-up states that Grok correctly identified three, which selects the former.

These counts are therefore \emph{derived}, not raw. They are exact under the stated arithmetic, but they do not restore per-scene labels: quantities that require scene-level alignment, such as the reported $85.7\%$ Gemini\textendash{}Grok agreement, remain unrecomputable and are reproduced from the original analysis.

\subsection{Results}
On materialized metaphor the three labellers diverge in a way no summary statistic captures. Against a human count of 9: the detector marked 72 scenes present (TP 7, FP 65, FN 2); Gemini marked exactly 1, and that one was correct; Grok marked none at all. Grok's $91\%$ raw agreement on this rule is achieved entirely by saying \textsc{absent} 100 times.

On atmosphere contradiction \textemdash{} the one rule with a usable distribution \textemdash{} the human marked 44 scenes. The detector marked 0, Gemini 2, Grok 6 (of which 3 were correct). All three missed the large majority of the positive class.

The two language models also diverge sharply from \emph{each other}. On micro-focus, Gemini marked 9 scenes and Grok marked 82 \textemdash{} against a human count of 96. Two systems applying the same written definition to the same hundred scenes produced answers an order of magnitude apart.

\section{Study 2b: Claude Fable 5 and ChatGPT 5.5}

\begin{table}[htbp]
\centering
\small
\resizebox{\textwidth}{!}{%
\begin{tabular}{lrrrrrrrrr}
\toprule
 & Human & \multicolumn{4}{c}{Claude Fable 5 (High)} & \multicolumn{4}{c}{ChatGPT 5.5} \\
\cmidrule(lr){3-6}\cmidrule(lr){7-10}
Rule & + & + & $\kappa$ & raw & TP/FP/FN & + & $\kappa$ & raw & TP/FP/FN \\
\midrule
Emotion label         &  0 &   0 & n/a & 100\% & 0/0/0 &  0 & n/a & 100\% & 0/0/0 \\
Simile                &  1 &   0 & \phantom{-}0.000 & 99\% & 0/0/1 &  0 & \phantom{-}0.000 & 99\% & 0/0/1 \\
Materialized metaphor &  9 &  78 & \phantom{-}0.027 & 29\% & 8/70/1 & 40 & \phantom{-}0.019 & 59\% & 4/36/5 \\
Micro-focus           & 96 & 100 & \phantom{-}0.000 & 96\% & 96/4/0 & 92 & \phantom{-}0.296 & 92\% & 90/2/6 \\
Temporal anchor       & 99 & 100 & \phantom{-}0.000 & 99\% & 99/1/0 & 98 & -0.014 & 97\% & 97/1/2 \\
Atmosphere contrad.   & 44 &  55 & \phantom{-}0.269 & 63\% & 31/24/13 & 42 & \phantom{-}0.184 & 60\% & 23/19/21 \\
\midrule
Overall raw           &    & \multicolumn{4}{c}{81.0\%} & \multicolumn{4}{c}{84.5\%} \\
\bottomrule
\end{tabular}%
}
\caption{Study 2b ($n=100$). Same scenes, same locked human reference, same prompt blocks as Study 2. All figures recomputed from published label files.}
\label{tab:study2b}
\end{table}

Two findings differ from Study 2 and one confirms it.

\textbf{Confirmed: materialized metaphor remains at chance.} Claude $\kappa = 0.027$, ChatGPT $\kappa = 0.019$. Claude marked 78 scenes present against the human's 9, producing $29\%$ raw agreement \textemdash{} the lowest single cell in all three studies. A model that agrees with a human on 29 of 100 binary judgements is not partially right; it is applying a different rule.

\textbf{New: atmosphere contradiction is partially recovered.} Claude $\kappa = 0.269$, ChatGPT $\kappa = 0.184$. Both are far from strong agreement, but both are clearly above chance, on the one rule whose class distribution can support that judgement. The Study 2 conclusion that machines cannot detect this feature does not generalise: it holds for the Study 2 raters and not for these two. We record this as a correction to the earlier write-up.

\textbf{New: raw agreement improves while $\kappa$ does not.} ChatGPT's $84.5\%$ overall is the highest figure in any study, and its materialized-metaphor $\kappa$ is $0.019$. The two facts are compatible because five of six rules have skewed human distributions \textemdash{} the first of the two well-known kappa paradoxes \cite{feinstein1990}. Any reader \textemdash{} or benchmark leaderboard \textemdash{} that ranks these systems by raw agreement will rank them by their willingness to say \textsc{absent}.

\section{Synthesis}

\subsection{One rule, five machines, five different answers}
The five machine labellers scored on materialized metaphor over the same 100 scenes returned:

\begin{center}
\resizebox{\textwidth}{!}{%
\begin{tabular}{lrrrrrr}
\toprule
 & Grok & Gemini 2.5 & ChatGPT 5.5 & Detector & Claude Fable 5 & \textbf{Human} \\
\midrule
Scenes marked present & 0 & 1 & 40 & 72 & 78 & \textbf{9} \\
\bottomrule
\end{tabular}%
}
\end{center}

The spread covers essentially the entire available range. One prior conjecture in this framework anticipates a directional failure of this kind \textemdash{} that language models represent a shown layer by substituting an abstract summary label for it \cite{bulut2026sb}. These data are consistent with that conjecture but do not test it: the conjecture predicts a direction, and the observed spread runs in both directions at once. Its registered experiment has not been run. This is not a ranking of accuracy; it is evidence that the five systems are not attempting the same task. A definition that yields $0$ from one competent reader and $78$ from another is not yet functioning as a definition.

\subsection{Chance-level agreement, repeated}
Across both human references and both disjoint scene sets, agreement on materialized metaphor was: detector vs.\ author $0.004$; detector vs.\ independent rater $0.015$; Grok $0.000$; ChatGPT $0.019$; Claude $0.027$. Gemini's $0.185$ is the single exception and rests on one positive judgement.

Five near-zero results, obtained from different labellers against different humans on different scenes, are unlikely to be an accident of any one of those factors.

\subsection{Two readings, and why we cannot choose}
\textbf{Reading (a): the feature is genuinely inferential.} Deciding whether a physical detail is carrying an unstated inner state requires reconstructing what the text withholds \textemdash{} an operation the framework locates on the reader's side rather than the text's \cite{bulut2026rpl, bulut2026twopath}, and whose predicted outcome it describes as statistical rather than deterministic \cite{bulut2026convergence}. If that is a reader-side inferential act rather than a text-side property, no surface detector should find it, and current language models \textemdash{} which the evidence suggests approach the task by pattern rather than by reconstruction \textemdash{} should not either. Under this reading the low agreement is a finding about machines.

\textbf{Reading (b): the definition is not operational.} The rule as written asks whether an abstract state has been rendered as concrete detail. In a corpus built entirely from that instruction, nearly every scene satisfies it under a permissive reading and few under a strict one. The human's 9-out-of-100 is itself a strict-reading artefact; the author's own drift in Study 1 is direct evidence that the criterion is unstable even within one rater. Under this reading the low agreement is a finding about the rule.

The two readings make one distinguishing prediction. Under (a), two independent human raters should agree with each other substantially better than any machine agrees with either. Under (b), two independent humans should also disagree. We note that the second outcome would not automatically condemn the rule: a body of work argues that human label variation on interpretive tasks is genuine signal about the task rather than noise to be minimised \cite{plank2022}. Distinguishing an unstable definition from a legitimately variable judgement would then become the next question rather than the closing one.

\textbf{We do not have that second human rater, and therefore we do not decide.} A reader who wishes to use the corpus's \texttt{applied\_rules} field should note that under both readings the field is unreliable for inferential features; the readings differ on where the fault lies, not on the practical conclusion.

\section{Limitations}

\subsection{The benchmark's own defect}
Five of the six rules have severely imbalanced human distributions. In Study 2, human positive counts were $0$, $1$, $9$, $96$, $99$ and $44$. On four of these, $\kappa$ carries almost no information and high raw agreement reflects the imbalance rather than any skill. Only atmosphere contradiction (44/56) sits in a usable range, and it is on that rule \textemdash{} and only that rule \textemdash{} that we are willing to say two models performed above chance.

The effect is well documented: $\kappa$ measures not only agreement but also the marginal distribution, so a coefficient reported without its prevalence and bias indices can mislead \cite{byrt1993}. This is a defect of the evaluation set, not of the systems evaluated. A corpus constructed so that nearly every scene has a temporal anchor cannot be used to measure whether anything detects temporal anchors. The design lesson for the next annotation batch is explicit: sample for balance on the target feature, or accept that the feature is untested.

\subsection{Single human rater per study}
Each study has exactly one human reference, and Study 1's is the scheme's own author. Neither human's labels have themselves been validated. Every number in this report is agreement \emph{with a particular person}, not accuracy.

\subsection{Provenance of specific figures}
Two caveats affect what can be independently recomputed.

\emph{The Gemini and Grok per-scene labels are not available.} Their confusion counts in Table \ref{tab:study2} were recovered arithmetically from the reported $\kappa$ and agreement values by the procedure in Section \ref{sec:recovery}, not read off label files. The recovery is exact for eleven of twelve cells and resolves an earlier internal record that gave Gemini's materialized-metaphor positive count as approximately 2: the correct figure is 1. Scene-level quantities remain unrecoverable, and the Gemini\textendash{}Grok agreement figure is reproduced rather than recomputed.

\emph{The detector's Study 2 labels are reconstructed.} They were derived from the \texttt{applied\_rules} field already published in the corpus, mapped through the published identifier mapping. The derivation script is published, and the reconstruction reproduces the original Study 2 detector column exactly. It is nonetheless a reconstruction.

\subsection{Interface conditions}
All model runs used public web interfaces rather than APIs, in fresh sessions without project context or memory. Temperature and system-prompt conditions were therefore not controlled. Model versions are named as reported by their providers at the time of running.

\subsection{Language and scope}
All scenes are Turkish and all were generated within one methodology. Whether these results transfer to other languages, to naturally occurring prose, or to other annotation schemes is untested.

\section{What would settle it}

Three steps, in order of value.

\textbf{A second independent human rater} on the same 100 scenes, blind to everything reported here. This is the one measurement that separates readings (a) and (b), and no further machine run substitutes for it.

\textbf{A balanced evaluation set} for the four working rules \textemdash{} scenes sampled so that each feature is present in roughly half. Without this, four of six rules remain untestable no matter how many systems are run.

A registered protocol for independent replication of the wider methodology already exists and has not yet been run by anyone \cite{bulut2026opct}; the reliability question addressed here is prior to it, since a protocol built on features that raters cannot apply consistently will not produce interpretable results.

\textbf{An operational rewrite of the materialized-metaphor rule}, with worked positive and negative examples at the boundary, followed by re-labelling. If two humans then agree and machines still do not, reading (a) is supported. If two humans still disagree, the rule needs replacing rather than detecting.

\section{Conclusion}

A rule-based detector and four large language models were scored against blind human labels on a six-feature annotation scheme, across three studies and two disjoint scene sets. Surface features \textemdash{} simile \textemdash{} were detected reliably. Inferential features were not. On materialized metaphor the five machine labellers marked between 0 and 78 of 100 scenes as positive against a human count of 9, and agreement was at chance for all but one, whose exception rests on a single judgement.

Whether this is a limit of the machines or a defect of the rule is not settled by these data, and this report declines to settle it. What the data do settle is narrower and still worth stating: the annotation layer shipped with this corpus should not be treated as ground truth for inferential features, and evaluations of such features that report raw agreement without class distributions will overstate performance substantially.

\section*{Data availability}

All human reference labels, model label files, detector labels, prompt blocks, identifier mapping and scoring scripts are published under CC BY-NC-ND 4.0 in the \texttt{evaluation/} directory of the Objective Projection dataset: \url{https://huggingface.co/datasets/leventbulut/objective-projection}. The results in Sections 3\textendash{}5 are reproduced by running the published scoring script against the published label files.

\section*{Acknowledgements}

The author thanks the volunteer rater who produced the Study 2 reference labels for accepting a tedious task with no stake in its outcome.

\end{document}